\documentclass{bmvc2k}
\usepackage[T1]{fontenc}
\usepackage[inline]{enumitem}
\usepackage{amsmath}
\usepackage{amssymb}
\usepackage{amsbsy}
\usepackage[separate-uncertainty=true,multi-part-units=single]{siunitx}
\usepackage{mathtools}
\usepackage{nicefrac}
\usepackage{bm}
\usepackage{threeparttable}
\usepackage{booktabs}
\usepackage{multirow}
\usepackage{tikz}
\usepackage{tikzscale}
\usepackage{standalone}
\usepackage{floatrow}
\usepackage{relsize}
\usepackage{graphicx}

\usetikzlibrary{calc}
\usetikzlibrary{positioning}
\usetikzlibrary{backgrounds}

\newlength\simheight
\definecolor{materialviolet}{RGB}{98, 0, 238}
\definecolor{materialteal}{RGB}{3, 218, 198}
\definecolor{darkgray178}{RGB}{178,178,178}
\definecolor{gainsboro229}{RGB}{229,229,229}
\definecolor{lightgray204}{RGB}{204,204,204}
\definecolor{mediumvioletred1771695}{RGB}{177,16,95}
\definecolor{bmvccaptioncolor}{RGB}{0,0,102}

\title{Visible Structure Retrieval\\for Lightweight Image-Based Relocalisation}

\addauthor{Fereidoon Zangeneh}{fereidoon.zangeneh@univrses.com}{1,2}
\addauthor{Leonard Bruns}{leonardb@kth.se}{1}
\addauthor{Amit Dekel}{amit.dekel@univrses.com}{2}
\addauthor{Alessandro Pieropan}{alessandro.pieropan@univrses.com}{2}
\addauthor{Patric Jensfelt}{patric@kth.se}{1}

\addinstitution{
 Division for Robotics,\\Perception and Learning (RPL)\\
 KTH Royal Institute of Technology\\
 Stockholm, Sweden
}
\addinstitution{
 Univrses AB\\
 Stockholm, Sweden
}

\runninghead{Zangeneh, Bruns, Dekel, Pieropan, Jensfelt}{Visible Structure Retrieval}

\DeclarePairedDelimiterX{\infdivx}[2]{\big(}{\big)}{%
  #1\;\delimsize\|\;#2%
}
\newcommand{\kldiv}{D_{\text{KL}}\infdivx}

\DeclareMathOperator{\E}{\mathbb{E}}

\begin{document}

\maketitle

\begin{abstract}
Accurate camera pose estimation from an image observation in a previously mapped environment is commonly done through structure-based methods: by finding correspondences between 2D keypoints on the image and 3D structure points in the map. In order to make this correspondence search tractable in large scenes, existing pipelines either rely on search heuristics, or perform image retrieval to reduce the search space by comparing the current image to a database of past observations. However, these approaches result in elaborate pipelines or storage requirements that grow with the number of past observations. In this work, we propose a new paradigm for making structure-based relocalisation tractable. Instead of relying on image retrieval or search heuristics, we learn a direct mapping from image observations to the visible scene structure in a compact neural network. Given a query image, a forward pass through our novel \textit{visible structure retrieval} network allows obtaining the subset of 3D structure points in the map that the image views, thus reducing the search space of 2D-3D correspondences. We show that our proposed method enables performing localisation with an accuracy comparable to the state of the art, while requiring lower computational and storage footprint.
\end{abstract}

\section{Introduction}
Camera relocalisation is the task of estimating the six-degree-of-freedom pose of a camera from what it views in a previously mapped environment. It is an essential component in autonomous outdoor operations in the absence of GPS, as well as augmented reality applications \cite{burki2019vizard}. Since the early days of relocalisation research \cite{se2005vision,schindler2007city,cummins2008fab}, improving accuracy and robustness has been the primary focus of different methods. These efforts culminated in solutions that at mapping time represent the scene in a 3D model of its salient structure points. They can then localise a query image by finding correspondences between an extracted set of 2D keypoints on the image and the 3D structure points in the map \cite{liu2017efficient, sattler2016efficient, sarlin2019coarse}. This structure-based paradigm has stood the test of time, such that its seminal works \cite{sattler2012improving, sattler2016efficient, sarlin2019coarse}, years after their publication, continue to reappear as strong baselines to benchmark the effectiveness of newly proposed localisation methods \cite{sarlin2021back, brachmann2021visual, wang2024glace}.

The robustness and high accuracy of the structure-based paradigm come with a pitfall: localisation becomes increasingly intractable as map size increases. To find 2D-3D correspondences between a query image and the map, a purely structure-based pipeline performs direct matching of the visual descriptors \cite{lowe1999object, detone2018superpoint} of the observed 2D keypoints and the mapped 3D points. Although such descriptors are locally discriminative, they suffer from perceptual aliasing in large environments. Specifically, while a simple nearest neighbour search of descriptors tends to find good matches in small search spaces, distinct points in a large map may be described by similar descriptors, leading to incorrect matches \cite{sattler2011fast}. Therefore, structure-based localisation is usually accompanied by a remedying strategy, such as using heuristics to assist the correspondence search \cite{sattler2012improving, sattler2016efficient}, or preceding it with image retrieval to leverage the global appearance information available in the image \cite{sarlin2019coarse}. Although proven effective, these approaches often involve complex processing pipelines or rely on explicitly storing past observations---a paradigm that incurs increasing storage and computational costs as the number of observations grows, regardless of their informational value for localisation.

\begin{figure}[t]
    \floatbox[{\capbeside\thisfloatsetup{capbesideposition={right,top},capbesidewidth=7.0cm}}]{figure}[\FBwidth]
{\caption{Our proposed method, visbile structure retrieval, retrieves the subset of SfM points that are visible per image observation. This lightweight setup serves to reduce the search space for establishing 2D-3D correspondences in a structure-based localisation pipeline, enabling fast and accurate localisation in large scenes.}\label{fig:teaser}}
{\scalebox{0.75}{\includegraphics[]{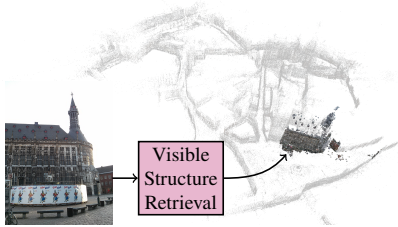}}}
\end{figure}

In this work, we revisit the mechanism used to make structure-based localisation tractable in large maps. We propose a novel setup that can effectively reduce the search space of 2D-3D correspondences for a query image in a map, leveraging a representation of the scene itself rather than explicitly storing observations of it. Such a representation ensures that the computational and storage footprint scales with the scene's complexity, rather than the volume of observations collected during mapping. We explore a new direction, applying the design elements of absolute pose regression \cite{kendall2015posenet} and scene coordinate regression \cite{brachmann2018learning} to this end. We show that at mapping time, a small scene-specific neural network can be optimised to regress regions of the scene structure that are visible per camera view. To localise an unseen query image, this network helps to efficiently reduce the search space of 2D-3D correspondences, such that direct nearest neighbour search of descriptors yields accurate pose estimation. This sidesteps the heuristic-based correspondence search or image retrieval and pairwise matching routines, while retaining the final geometry-based robust estimation of camera pose in a structure-based pipeline. We show that this \textit{visible structure retrieval} network can be formulated as the decoder of a variational autoencoder pipeline that, conditioned on an image, is trained to reconstruct its triangulated 3D points in the map.

In summary:
\begin{enumerate*} [label=(\arabic*)]
    \item We introduce visible structure retrieval, a novel paradigm for scaling structure-based localisation to large scenes, by directly retrieving structure points as seen by a query image.
    \item We propose learning the visible structure retrieval operation in a small scene-specific network, and formulate it as a generative modelling task.
    \item We show that this network can be trained in a variational framework without requiring any supervision beyond what is available from Structure-from-Motion.
    \item We perform thorough evaluation to show that our method enables localisation with an accuracy comparable to the state of the art, while requiring a lower computational and storage footprint.
\end{enumerate*}

\section{Related Work}

Our work serves as a preamble to structure-based localisation, replacing image retrieval in a hierarchical setup; it incorporates design elements from absolute pose regression as well as scene coordinate regression. We now briefly review each of these fields.

\textbf{Structure-based methods} in camera relocalisation comprise approaches that explicitly model the scene structure, typically by a set of 3D points. At mapping time, the scene is modelled by registering the mapping images through Structure-from-Motion (SfM) \cite{schonberger2016structure, pan2024global}, matching and triangulating their 2D keypoints to create a 3D model. To localise a query image, its 2D keypoints are extracted and, through their local descriptors, exhaustively matched to the 3D points in the map. From the possibly noisy set of matches, the camera pose is then computed with a Perspective-n-Point (PnP) \cite{kneip2011novel} procedure in a robust estimation loop \cite{sattler2011fast}. Given sufficient inliers, these methods make accurate predictions. However, local descriptors suffer from perceptual aliasing as the map size increases. This leads to high noise levels in the set of matches, from which robust estimation may not recover. In other words, structure-based localisation can become intractable in large maps. A prominent mitigating approach is to follow a heuristic of using visual vocabulary trees to assess the discriminativeness of descriptors, together with covisibility assumptions to speed up the correspondence search \cite{sattler2012improving, sattler2016efficient}. However, this process can still inhibit applications with low computing resources. 

\textbf{Image retrieval} refers to modelling a scene by a database of representative images with known poses, so that given a query, the most similar database image can be retrieved to produce a coarse estimate of the camera pose \cite{torii201524}. To efficiently perform this database search, each image is summarised by a global descriptor vector \cite{jegou2010aggregating, arandjelovic2016netvlad, zhu2023r2former, keetha2023anyloc}. Structure-based localisation can benefit from image retrieval within a hierarchical framework \cite{sarlin2019coarse}, where matching the top-ranked reference images for a given query effectively constrains the 2D–3D correspondence search space. This setup also enables the use of powerful learned feature matchers \cite{sarlin2020superglue, lindenberger2023lightglue}. However, pairwise matching of reference images introduces additional computational and memory overhead in the localisation pipeline.

\textbf{Absolute pose regression} (APR) offers a memory-efficient solution to relocalisation, albeit fundamentally different in nature from traditional methods. While the latter group explicitly model the scene, APR methods implicitly encode the camera poses and the appearance they observe in the weights of a scene-specific neural network. At mapping time, the network is trained to directly regress the pose of the camera for each image it views \cite{kendall2015posenet}. This network is then trusted to, with a small memory footprint, perform fast inference of camera poses for unseen query images. The attractive test-time properties and the ease of deployment of APR networks promoted efforts to improve their training with more effective loss functions \cite{brahmbhatt2018geometry, chen2021direct, kendall2017geometric, zangeneh2023vapor} and network setups \cite{melekhov2017image, wang2020atloc, shavit2021learning}. This paradigm also enables relocalisation to be formulated as a generative modelling problem: localising a query image can be framed as inferring the posterior distribution of camera poses given that image. This can be learned in a variational autoencoder (VAE) framework, as shown in the context of localisation under observation ambiguities, where the posterior distribution may be multimodal \cite{zangeneh2023vapor}. Despite these efforts, APR is understood to exhibit retrieval-like behaviour. That is, it returns a pose similar to that of the closest training image. In other words, it will not generalise beyond the training images and may poorly interpolate between them \cite{sattler2019understanding}.

\textbf{Scene coordinate regression} (SCR) avoids the generalisation problem of APR by regressing a domain that remains stationary between mapping and localisation time: 3D coordinates of the scene structure. Instead of a single camera pose, the network is trained to predict the 3D coordinate of the observed scene structure per image patch on a dense grid \cite{brachmann2018learning}. This produces a set of 2D-3D correspondences, on which a robust estimation routine of PnP can estimate the camera pose. While originally proposed for RGB-D images \cite{shotton2013scene}, SCR has been extended to RGB-only supervision at mapping time \cite{brachmann2021visual}, and has been shown to achieve fast and accurate localisation with a small memory footprint \cite{brachmann2023accelerated}. However, this success has been limited to small scenes for SCR networks in their standard formulation. The convergence of such networks relies on learning local patch-level features that are discriminative enough so that the observed 3D structure can be detected, while remaining invariant enough so that view changes do not affect the predicted 3D coordinates. Due to this local nature of the predictions, perceptual aliasing in large-scale scenes perplexes the patch features and prevents the SCR network from converging to the correct geometry. Remedies to this include splitting large scenes and training multiple networks \cite{brachmann2023accelerated}, providing additional global context \cite{wang2024glace}, or learning a covisibility graph and getting assistance from image retrieval at test time \cite{jiang2025r}.

\textbf{In this work}, equipped with the recent advances in APR and SCR, we revisit the challenge of structure-based localisation in large-scale scenes. While retaining the final structure-based pose estimation procedure for its accuracy, we replace the preceding heuristic-based correspondence search \cite{sattler2016efficient} or image retrieval \cite{sarlin2019coarse} with our regression-based visible structure retrieval method for its fast inference, small memory footprint, and simplicity. We take inspiration from APR to formulate coarse relocalisation as a generative modelling problem conditioned on global image features \cite{zangeneh2024cvaepor} for its scalability to large scenes; at the same time, similar to SCR, we opt to predict scene structure coordinates for their robustness to domain shift between mapping and localisation time \cite{brachmann2023accelerated}.

\section{Method: Visible Structure Retrieval}
We propose a method that, given a query image, retrieves the subset of 3D structure points in the SfM map that are visible from that view, as illustrated in Fig.\ \ref{fig:teaser}. In a structure-based localisation pipeline, our method efficiently reduces the search space of establishing 2D-3D correspondences to a size where a nearest neighbour search of descriptors proves effective. This obviates the need for advanced search strategies or image retrieval. We pose visible structure retrieval as learning the mapping from image features to the set of 3D points it observes. We propose formulating this mapping as a generative model in Section \ref{sec:generation} and show how it can be learned in Section \ref{sec:training}. We then show how the predictions of the learned model can be used to retrieve the original 3D points from the SfM map in Section \ref{sec:lookup}. Section \ref{sec:pipeline} details the integration of our method as part of a lightweight relocalisation pipeline.

\subsection{Generative modelling of structure regression}
\label{sec:generation}
In structure-based relocalisation, keypoints on a query image $\boldsymbol{x} \in \mathbb{R}^{H \times W \times 3}$ are exhaustively matched to the structure points $\boldsymbol{y} \in \mathbb{R}^3$ in the SfM map by a nearest neighbour search of their descriptors. We are interested in limiting this search space to the posterior distribution $p(\boldsymbol{y} \mid \boldsymbol{x})$ of structure points $\boldsymbol{y} \in \mathbb{R}^3$ that are visible in the image. This posterior distribution over points can have arbitrary shapes and span arbitrary regions in $\mathbb{R}^3$ depending on the camera pose. Moreover, the sparsity of the SfM map often leads to non-uniform point densities across the structure manifold. Therefore, the learning of $p(\boldsymbol{y} \mid \boldsymbol{x})$ has to be robust to such variations. We propose modelling  $p(\boldsymbol{y} \mid \boldsymbol{x})$ at mapping time, by training a small scene-specific neural network $f_\theta(\cdot)$ that, given an input image $\boldsymbol{x}$ and a noise sample $\boldsymbol{z} \sim \mathcal{N}(\mathbf{0}, \mathbf{I})$, generates a sample $\boldsymbol{y}$ from the posterior distribution of 3D points on the visible structure. As such, an arbitrarily large number of noise samples, $N$, can be drawn $\mathcal{Z} = \{\boldsymbol{z}_i \sim \mathcal{N}(\mathbf{0}, \mathbf{I})) \mid i=1:N\}$ to generate a representative set of structure points $\mathcal{Y} = \{\boldsymbol{y}_i = f_\theta(\boldsymbol{z}_i, \boldsymbol{x}) \mid \boldsymbol{z}_i \in \mathcal{Z}\}$ for any image $\boldsymbol{x}$. This is illustrated in Fig.\ \ref{fig:vae}b. This formulation implies that $f_\theta(\cdot)$ should learn a random variable transformation between $\boldsymbol{z} \in \mathbb{R}^d$ and $\boldsymbol{y} \in \mathbb{R}^3$ per image observation $\boldsymbol{x}$, transforming the densities of $p(\boldsymbol{z}) = \mathcal{N}(\mathbf{0}, \mathbf{I})$ to $p(\boldsymbol{y} \mid \boldsymbol{x})$. In other words, the space of $\boldsymbol{z}$ can be interpreted as the latent space of the visible scene structure. This organisation of the latent space according to the chosen prior $p(\boldsymbol{z})$ can be learned in a VAE setup.

\subsection{Training in a variational autoencoder pipeline}\label{sec:training}

\begin{figure}[t]
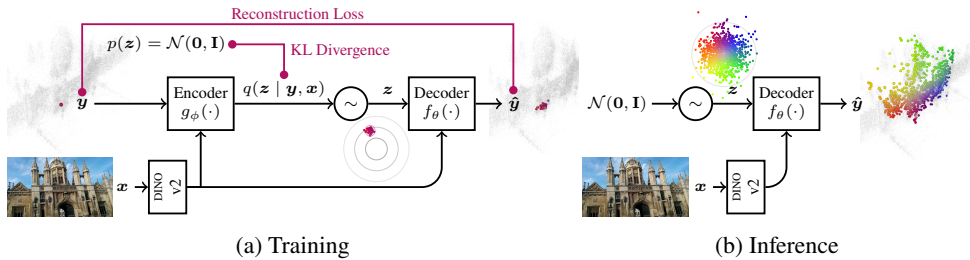

    \centering
    \begin{subfigure}{0.59\linewidth}%
        \centering%
        \scalebox{0.87}{\includegraphics[]{images/2_vistr_training.tikz}}
        \caption{Training}%
    \end{subfigure}%
    \begin{subfigure}{0.40\linewidth}%
        \centering%
        \scalebox{0.87}{\includegraphics[]{images/2_vistr_inference.tikz}}
        \caption{Inference}%
    \end{subfigure}%
    \caption{(a) Our visible structure retrieval network is trained as the decoder of a VAE pipeline that, for an image observation $\boldsymbol{x}$, reconstructs its visible 3D structure points $\boldsymbol{y} \in \mathbb{R}^3$. Specifically, given image-level features of an observation $\boldsymbol{x}$, each structure point $\boldsymbol{y}$ from the SfM map that is visible in that image is encoded to its unique latent posterior $q(\boldsymbol{z} \mid \boldsymbol{x}, \boldsymbol{y})$, while the decoder is tasked with decoding latent samples $\boldsymbol{z} \in \mathbb{R}^d$ from this posterior back to the original point. At the same time, latent posteriors of different points visible in $\boldsymbol{x}$ are constrained to collectively conform to the prior $p(\boldsymbol{z}) = \mathcal{N}(\mathbf{0}, \mathbf{1})$. This training scheme organises the latent space so that it can be interpreted as the space of visible structure per image observation $\boldsymbol{x}$. (b) At inference time, the decoder maps noise samples from the prior distribution $\mathcal{N}(\mathbf{0}, \mathbf{1})$ to different regions of the scene structure visible in the observed image.}
    \label{fig:vae}
\end{figure}

We propose to train the network $f_\theta(\cdot)$ that models $p(\boldsymbol{y} \mid \boldsymbol{x})$ as the decoder of a conditional VAE pipeline with latent prior distribution $p(\boldsymbol{z}) = \mathcal{N}(\mathbf{0}, \mathbf{I})$. This autoencoder pipeline, with encoder $g_\phi(\cdot)$ and decoder $f_\theta(\cdot)$, is trained to reconstruct the visible SfM points conditioned on each mapping image, as shown in Fig.\ \ref{fig:vae}a.

\textbf{Setting:} Given the SfM map for a scene, we gather the set of triangulated points $\{\boldsymbol{y}_{k,l} \in \mathbb{R}^{3}\}_{l=1:L_k}$ per mapping image $\boldsymbol{x}_k \in \mathbb{R}^{H\times W\times3}$. We then create a shuffled dataset of point-image pairs $\mathcal{D}_\text{train} = \{(\boldsymbol{x}_n, \boldsymbol{y}_n)\}$, where each image is repeated in as many pairs as it has triangulated points in the map. The VAE then reconstructs these point samples conditioned on their image observations. Specifically, the encoder, conditioned on the image $\boldsymbol{x}_n$, encodes the structure point $\boldsymbol{y}_n$ to its inferred latent posterior distribution $p(\boldsymbol{z} \mid \boldsymbol{y}_n, \boldsymbol{x}_n)$, modelled as Gaussian by its mean and covariance. The decoder, also conditioned on $\boldsymbol{x}_n$, then maps back samples drawn from this posterior $\boldsymbol{z}_j \sim p(\boldsymbol{z} \mid \boldsymbol{y}_n, \boldsymbol{x}_n)$ to reconstructions $\hat{\boldsymbol{y}}_{n,j} \in \mathbb{R}^3$ of $\boldsymbol{y}_n$.

\textbf{Optimisation:} The conditional VAE is trained by maximising the evidence lower-bound (ELBO) \cite{kingma2019introduction, sohn2015learning}, which in turn maximises the conditional likelihood, derived as
\begin{equation} \label{eq:elbo}
    \begin{split}
    &\log p(\boldsymbol{y} | \boldsymbol{x}) - \overbrace{\kldiv{q_\phi(\boldsymbol{z} \mid \boldsymbol{y}, \boldsymbol{x})}{p(\boldsymbol{z} \mid \boldsymbol{y}, \boldsymbol{x})}}^{\geq 0} \\
    &= \underbrace{\E_{q_\phi(\boldsymbol{z} \mid \boldsymbol{y}, \boldsymbol{x})} \log p_\theta(\boldsymbol{y} \mid \boldsymbol{z}, \boldsymbol{x}) - \kldiv{q_\phi(\boldsymbol{z} \mid \boldsymbol{y}, \boldsymbol{x})}{p(\boldsymbol{z})}}_{\text{ELBO}},
    \end{split}
\end{equation}
where $q_\phi(\boldsymbol{z} \mid \boldsymbol{y}, \boldsymbol{x})$ denotes the inferred latent posterior by the encoder $g_\phi(\cdot)$, and $p_\theta(y \mid \boldsymbol{z}, \boldsymbol{x})$ denotes reconstruction likelihood through the decoder $f_\theta(\cdot)$. We compute the expected value of the latter with Monte Carlo samples from $q_\phi(\boldsymbol{z} \mid \boldsymbol{y}, \boldsymbol{x})$, following a Gaussian model
\begin{equation} \label{eq:reconstruction}
        \log p_\theta\big(\boldsymbol{y} \mid \boldsymbol{z}, \boldsymbol{x}\big) = -\nicefrac{1}{2}\big(3\log 2\pi + \log \det(\bm{\Sigma}) + (\boldsymbol{y} - \hat{\boldsymbol{y}})^T \bm{\Sigma}^{-1} (\boldsymbol{y} - \hat{\boldsymbol{y}}) \big),
\end{equation}
where a learnable $3\times3$ covariance matrix $\bm{\Sigma}$ is optimised alongside network parameters $\theta$ and $\phi$ to automatically adjust the importance weight of reconstruction error throughout training.

\subsection{Retrieval of the regressed structure}
\label{sec:lookup}
Once trained, we can easily feed samples from the standard Gaussian prior distribution $\boldsymbol{z} \sim \mathcal{N}(\mathbf{0}, \mathbf{I})$ together with a conditioning image $\boldsymbol{x}$ through the decoder $f_\theta(\cdot)$, to simulate its posterior distribution of visible structure points $p(\boldsymbol{y} \mid \boldsymbol{x})$. This yields a generated point set $\hat{\mathcal{Y}} = \{\hat{\boldsymbol{y}}_i = f_\theta(\boldsymbol{z}_i, \boldsymbol{x}) \mid \boldsymbol{z}_i \sim \mathcal{N}(\mathbf{0}, \mathbf{I}), i=1:N\}$. However, these samples are not immediately usable for correspondence search: the regression loss can merely ensure that the generated points lie on the learned structure manifold, but it does not perform selection from the set of sparse SfM points. To identify the subset of original SfM points that are visible, we simply select all SfM points in the vicinity of the generated ones. This can be efficiently performed through a k-d tree search of SfM points $\mathcal{P}_\text{SfM}$ within a radius $r$ of the generated points: $\tilde{\mathcal{Y}}_r = \{\tilde{\boldsymbol{y}} \in \mathcal{P}_\text{SfM} \subset \mathbb{R}^3 \mid \exists\, \hat{\boldsymbol{y}} \in \hat{\mathcal{Y}} \text{ s.t. } \|\hat{\boldsymbol{y}} - \tilde{\boldsymbol{y}}\| \leq r\}$. Each of these retrieved 3D SfM points is associated with a descriptor that enables 2D–3D matching. To ensure an efficient radius search on the k-d tree, we propose voxel downsampling of the generated points. This also mitigates the tendency of the VAE’s Gaussian prior to generate densely clustered samples near the centre of the data space.

\section{Lightweight Image-Based Relocalisation}
\label{sec:pipeline}
We now lay down the steps required to perform relocalisation using visible structure retrieval.

\textbf{Mapping} of the scene begins similarly to existing structure-based approaches \cite{sattler2016efficient, sarlin2019coarse, sarlin2021back}, using an SfM procedure on a collection of mapping images to obtain 3D structure points and their associated descriptors. However, our method differs in the subsequent stages. Traditional descriptor search heuristics rely on quantisation techniques \cite{sattler2016efficient}, which require non-trivial choices regarding the size and resolution of the visual vocabulary. Likewise, image retrieval approaches proceed by computing global image descriptors \cite{sarlin2019coarse}, which could pose a challenge in selecting a representative subset of reference images, particularly when the image collection is dense. Our method instead optimises a small neural network, without distinction, on all available reference images to reconstruct all mapped 3D points. The training procedure, outlined in Section \ref{sec:training}, replaces manual decisions with data-driven learning. To keep the scene-specific network compact and focused solely on learning the scene structure, we train it using image embeddings extracted by a general feature extractor---specifically, the base model of DINOv2 \cite{oquab2023dinov2} for its ability to capture fine-grained semantic information.

\textbf{Localisation} of a query image is done in four steps:
\begin{enumerate*} [label=(\arabic*)]
    \item the visible structure in the image is regressed from noise samples through a forward pass of our network;
    \item the mapped SfM points are retrieved through k-d tree radius search around the regressed points;
    \item keypoints of the query image are extracted with SuperPoint \cite{detone2018superpoint} and matched with the retrieved 3D points through nearest neighbour search in descriptor space; and
    \item the camera pose is estimated through PnP within a RANSAC framework.
\end{enumerate*} Our localisation pipeline, illustrated in Fig.\ \ref{fig:localisation}, sidesteps using search heuristics \cite{sattler2016efficient} or image retrieval and pairwise matching of images \cite{sarlin2019coarse}. Moreover, in contrast to task-specific descriptor models for image retrieval \cite{arandjelovic2016netvlad}, our use of DINOv2 leverages general-purpose features that can be reused by other tasks beyond localisation at run time, without incurring additional costs.

\begin{figure}[t]
    \centering
    \includegraphics[]{images/3_localization.tikz}%
    \caption{The novel \protect\tikz[baseline]{\protect\node[rounded corners=0pt, fill=mediumvioletred1771695!30, inner sep=2pt,anchor=base, draw=black, text=black] (A) {visible structure retrieval network};} is at the heart of our localisation pipeline: given image-level features of a query $\boldsymbol{x}$, it predicts the posterior distribution over visible structure points $p(\boldsymbol{y} \mid \boldsymbol{x})$. This is done through a forward pass of an arbitrarily large set of noise samples together with the image-level features of $\boldsymbol{x}$ to get a set of regressed 3D points $\hat{\mathcal{Y}}$. A radius search in the SfM point cloud's k-d tree around elements of $\hat{\mathcal{Y}}$ then retrieves a submap $\tilde{\mathcal{Y}}$ from the full SfM map. This effectively confines the search space of 2D-3D matches in large scenes, such that nearest neighbour matching of descriptors for query keypoints and $\tilde{\mathcal{Y}}$ points yields sufficient inliers for PnP to recover the query camera pose.}
    \label{fig:localisation}
\end{figure}

\section{Implementation Details} 
We implement the encoder and decoder networks as multilayer perceptrons, each with 5 layers of 512 LeakyReLU-activated neurons. The network input is a concatenation of image embeddings and a $64$-dimensional representation of either the 3D point---for the encoder, or the latent sample---for the decoder, obtained through another learned layer. We also draw a residual connection from the input to the third layer. The image embedding is the $768$-dimensional class token extracted by a pretrained base DINOv2 model \cite{oquab2023dinov2}. We opt for a $4$-dimensional latent space in all experiments, except for visualisations in the paper, where a $2$-dimensional latent space is used. The $3 \times 3$ output noise covariance matrix $\bm{\Sigma}$ in \eqref{eq:reconstruction} is parameterised by its lower triangular component from Cholesky decomposition. We train the networks for $100\mathrm{k}$ iterations with Adam optimiser and once-cycle scheduling of maximum learning rate $0.001$ \cite{smith2019super}. We use a batch size of $128$ images and $50$ random SfM points each, and draw $50$ Monte Carlo samples to estimate ELBO's expected reconstruction likelihood term in \eqref{eq:elbo}. We found a cyclical $0 \rightarrow 1$ warm-up of KL divergence term after $20\mathrm{k}$ iterations and with a period of $2\mathrm{k}$ iterations to help with better convergence \cite{fu2019cyclical}. We perform data augmentation by adding Gaussian noise with $\sigma^2=1$ to the training DINOv2 embeddings. For numerical stability, we constrain the VAE to reconstruct 3D points within the normalised cube $[0,1]^3$. The likelihood \eqref{eq:reconstruction} is computed in this normalised space, and the learnable covariance matrix $\bm{\Sigma}$ is initialised at the start of training as a diagonal matrix with entries set to $0.1$. At inference, the reconstructed points are affinely mapped from the normalised cube back to the corresponding SfM coordinate range of each scene. We use $1000$ generated samples and $r=5$m for k-d tree radius search in all localisation experiments.


\section{Experiments}
We aim to evaluate the effectiveness and efficiency of visible structure retrieval in structure-based relocalisation. To this end, we design our evaluation protocol to measure
\begin{enumerate*} [label=(\arabic*)]
    \item localisation accuracy,
    \item storage footprint, and
    \item time to localise a query.
\end{enumerate*}

\textbf{Datasets} that we use for evaluation are the large-scale scenes of Cambridge Landmarks \cite{kendall2015posenet} and the city-scale Aachen-Day-Night \cite{sattler2018benchmarking}. The former contains mapping and query images from five outdoor locations in Cambridge, while the latter features daytime mapping images from Aachen’s old town, along with both daytime and nighttime query images that present challenging illumination and viewpoint variations. To evaluate localisation accuracy, we adopt the standard metrics used in prior work for each dataset: the median translation and rotation error, and localisation accuracy measured by recall at varying error thresholds.

\textbf{Baseline methods} that we benchmark our method against include HLoc (SP+SG) \cite{sarlin2019coarse} as the gold standard method in localisation literature---relying on image retrieval and 2D-2D pairwise matching of images, and Active Search \cite{sattler2016efficient}, a commonly referenced structure-based method. We also implement two other structure-based baseline variants, closer to our method, in which image retrieval is used to look up the subset of SfM points for direct 2D-3D matching using SuperPoint \cite{detone2018superpoint}. We use NetVLAD \cite{arandjelovic2016netvlad} and AnyLoc \cite{keetha2023anyloc} for this, and retrieve top-10 and 50 images for Cambridge Landmarks and Aachen queries, respectively. We also report the performance of top performing APR and SCR representatives, MS-Transformer \cite{shavit2021learning} and GLACE \cite{wang2024glace}, as alternative paradigms that learn an implicit representation of the scene rather than explicitly storing observations of it.

\section{Results and Discussion}

\begin{figure}[t]
    \centering
    \resizebox{12.8cm}{!}{\includegraphics[]{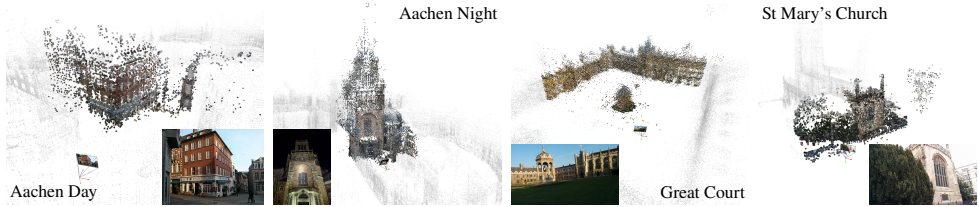}}%
    \caption{Our approach effectively retrieves the set of 3D points observed in a query image.}
    \label{fig:results}
\end{figure}

\begin{table}[t]
    \centering
    \begin{threeparttable}
        \caption{Median prediction error on scenes from Cambridge Landmarks (lower is better)}\label{tab:cambridge-error}
        \scriptsize
        \setlength{\tabcolsep}{1.7pt}
        \sisetup{detect-weight=true,detect-inline-weight=math}
        \begin{tabular}{@{}lcSScSScSScSScSScc@{}}
            \toprule
            \multirow{2.5}{*}{Method} & \multicolumn{2}{c}{King's College} & & \multicolumn{2}{c}{Old Hospital} & & \multicolumn{2}{c}{Shop Façade} & & \multicolumn{2}{c}{St Mary's Ch.} & & \multicolumn{2}{c}{Great Court} & & \multicolumn{2}{c}{Storage for Retrieval}\\
            \cmidrule{2-3}\cmidrule{5-6}\cmidrule{8-9}\cmidrule{11-12}\cmidrule{14-15}\cmidrule{17-18}
            & t~/~cm & R~/~$^\circ$ && t~/~cm & R~/~$^\circ$ && t~/~cm & R~/~$^\circ$ && t~/~cm & R~/~$^\circ$ && t~/~cm & R~/~$^\circ$ && Size / MB & Complexity \\
            \midrule
            HLoc (SP+SG) \cite{sarlin2019coarse} & $12$ & 0.2 && $15$ & 0.3 && $4$ & 0.2 && $7$ & 0.2 && $16$ & 0.1 && $44$ & $O(n)$\\
            Active Search \cite{sattler2016efficient} & $13$ & 0.2 && $20$ & 0.4 && $4$ & 0.2 && $8$ & 0.3 && $24$ & 0.1 && -- & --\\
            NetVLAD \cite{arandjelovic2016netvlad} (SP+NN) & $12$ & 0.2 && $15$ & 0.3 && $5$ & 0.2 && $8$ & 0.3 && $22$ & 0.1 && $44$ & $O(n)$\\
            AnyLoc \cite{keetha2023anyloc} (SP+NN) & $12$ & 0.2 && $16$ & 0.3 && $5$ & 0.2 && $7$ & 0.2 && $22$ & 0.1 && $527$ & $O(n)$\\
            \midrule
            MS-Transformer \cite{shavit2021learning} & $83$ & 1.5 && $181$ & 2.4 && $86$ & 3.1 && $162$ & 4.0 && \multicolumn{2}{c}{Not Converged} && $72$ & $O(1)$\\
            GLACE \cite{wang2024glace} & $19$ & 0.3 && $17$ & 0.4 && $4$ & 0.2 && $9$ & 0.3 && $19$ & 0.1 && $5\times9$  & $O(1)$\\
            Ours: ViStR (SP+NN) & $13$ & 0.2 && $16$ & 0.3 && $5$ & 0.2 && $8$ & 0.3 && $26$ & 0.1 && $5\times7$  & $O(1)$\\
            \bottomrule
        \end{tabular}
    \end{threeparttable}
\end{table}

\begin{table}[t]
    \centering
    \begin{threeparttable}
        \caption{Percentage of correctly localised query images from Aachen (higher is better)}\label{tab:aachen-error}
        \scriptsize
        \setlength{\tabcolsep}{3.1pt}
        \sisetup{detect-weight=true,detect-inline-weight=math}
        \begin{tabular}{@{}lSSSSSSSScc@{}}
            \toprule
            \multirow{2.5}{*}{Method} & \multicolumn{3}{c}{Aachen Day} & & \multicolumn{3}{c}{Aachen Night} & & \multicolumn{2}{c}{Storage for Retrieval}\\
            \cmidrule{2-4}\cmidrule{6-8}\cmidrule{10-11}
            & \multicolumn{1}{c}{0.25m~/~$2^\circ$} & \multicolumn{1}{c}{0.5m~/~$5^\circ$} & \multicolumn{1}{c}{5m~/~$10^\circ$} && \multicolumn{1}{c}{0.25m~/~$2^\circ$} & \multicolumn{1}{c}{0.5m~/~$5^\circ$} & \multicolumn{1}{c}{5m~/~$10^\circ$} && \multicolumn{1}{c}{Size / MB} & \multicolumn{1}{c}{Complexity}\\
            \midrule
            
            HLoc (SP+SG) \cite{sarlin2019coarse} & 89.8 & 96.1 & 99.4 && 77.0 & 90.6 & 100.0 && $56$ & $O(n)$\\ 
            Active Search \cite{sattler2016efficient} & 85.3 & 92.2 & 97.9 && 39.8 & 49.0 & 64.3 && -- & --\\  
            NetVLAD \cite{arandjelovic2016netvlad} (SP+NN) & 85.2 & 92.7 & 97.6 && 67.0 & 83.2 & 92.7 && $56$ & $O(n)$\\
            AnyLoc \cite{keetha2023anyloc} (SP+NN) & 85.2 & 92.8 & 98.1 && 67.5 & 82.7 & 96.3 && $658$ & $O(n)$\\
            \midrule
            MS-Transformer \cite{shavit2021learning} & 0.0 & 0.0 & 0.0 && 0.0 & 0.0 & 0.0 && $72$ & $O(1)$\\
            GLACE \cite{wang2024glace} & 8.6 & 20.8 & 64.0 && 1.0 & 1.0 & 17.3 && $27$ & $O(1)$\\
            Ours: ViStR (SP+NN) & 82.3 & 90.7 & 95.6 && 64.4 & 77.5 & 89.0 && $7$ & $O(1)$\\
            \bottomrule
        \end{tabular}
    \end{threeparttable}
\end{table}

\textbf{Effectiveness:} Table\ \ref{tab:cambridge-error} reports the median translation and rotation error of predictions for Cambridge Landmarks. We see that the accuracy of predictions made following our visible structure retrieval method is comparable to that of the state-of-the-art structure-based baselines and significantly better than the APR representative, MS-Transformer. This is noteworthy, as our method at its heart was inspired by APR, performing regression from global image features. We can see that our adaptation of this paradigm to regress structure points instead of camera poses addresses the generalisation shortcoming of APR, such that it can now be used as an effective initialisation step for structure-based localisation. Table\ \ref{tab:aachen-error} enables a similar analysis, presenting the localisation accuracy for Aachen query images. Our method outperforms the purely structure-based approach, Active Search, and achieves performance comparable to direct 2D-3D matching baselines that use NetVLAD or Anyloc. Although HLoc yields the best performance, it incurs a substantially higher computational cost, as we will discuss in the next section. Looking at the regression-based baselines, both MS-Transformer and GLACE perform poorly on this city-scale dataset: the APR-based method suffers from catastrophic failure in generalising from the training to the test domain, while the local patch-wise predictions in SCR are severely degraded by perceptual aliasing. This highlights the advantage of our generative formulation, which predicts structure points by leveraging global image embeddings rather than relying on local patches.

\textbf{Efficiency:} Table\ \ref{tab:cambridge-error} and \ref{tab:aachen-error} also report the storage requirements specifically for the retrieval stage of the localisation pipeline, that is, the stage to reduce the search space of correspondences. We can see that the storage requirements for approaches that rely on image retrieval grows linearly with the number of stored observations in the reference database---they have a complexity of $O(n)$, where $n$ is the number of observations. On the other hand, for implicit approaches---APR, SCR, and our method, the storage requirement does not depend on the number of past observations. For an implicit representation, the storage requirement reflects the number of network parameters required to model the scene. Table\ \ref{tab:aachen-time} reports the average time taken to localise an image using structure-based methods, and the share of each component in the pipeline. The times are measured on a PC with Intel Core i9-13900KF CPU and NVIDIA RTX 4090 GPU. Our pipeline can localise a query image in under $100$ms, which is $50$ times faster than HLoc, the top performing method in Table\ \ref{tab:aachen-error}. The time required for global search corresponds to the image retrieval operation in the baseline methods, and to a network forward pass in ours. We note that while image retrieval has linear complexity with respect to the size of the reference image database, our method achieves constant-time performance. As an implication of this, when reference images are densely collected during mapping, an image retrieval pipeline must resort to heuristic subsampling to remain scalable, whereas our method bypasses this challenge by learning a compact representation.

\begin{table}[t]
    \centering
    \begin{threeparttable}
        \caption{Average time in milliseconds to localise a daytime image from Aachen dataset
    \vspace{-0.5em}}\label{tab:aachen-time}
        \scriptsize
        \setlength{\tabcolsep}{6pt}
        \sisetup{detect-weight=true,detect-inline-weight=math}
        \begin{tabular}{@{}lcccccccc@{}}
            \toprule
            \multirow{3}{*}{Method} & Global & \multicolumn{2}{c}{\multirow{2}{*}{Global Search}} & SfM & Local & Local & \multirow{2}{*}{PnP} & \multirow{2}{*}{Total} \\
            & Feature & & & K-D Tree & Feature & Feature & \multirow{2}{*}{RANSAC} & \multirow{2}{*}{Time} \\
            & Extraction & Time & Complexity & Lookup & Extraction & Matching & & \\
            \midrule
            HLoc (SP+SG) \cite{sarlin2019coarse} & $19$ & $6$ & $O(n)$ & -- & $17$ & $5\times105$ & $26$ & $5318$\\
            Active Search \cite{sattler2016efficient} & -- & -- & -- & -- & $114$ & \multicolumn{2}{c}{---------~$112$\tnote{*}~---------} & $226$\\
            NetVLAD \cite{arandjelovic2016netvlad} (SP+NN) & $19$ & $6$ & $O(n)$ & -- & $17$ & $27$ & $22$ & $91$\\
            AnyLoc \cite{keetha2023anyloc} (SP+NN) & $45$ & $6$ & $O(n)$ & -- & $17$ & $25$ & $20$ & $113$ \\
            Ours: ViStR (SP+NN) & $4$ & $1$ & $O(1)$ & $8$ & $17$ & $33$ & $26$ & $89$ \\
            \bottomrule
        \end{tabular}
        \begin{tablenotes}
          \scriptsize
          \item[*] Time shared between matching and pose estimation, taken from \cite{sarlin2019coarse}.
        \end{tablenotes}
    \end{threeparttable}
\end{table}

\textbf{Future work:} While visible structure retrieval provides an effective and efficient alternative to image retrieval, its advantages may be overshadowed by the storage demands of a full relocalisation pipeline. In structure-based localisation, the retrieved 3D points must be matched to keypoints in the query image: typically through descriptor matching, as in our experiments. However, storing descriptors for all SfM points can require hundreds of megabytes or even gigabytes of memory. A promising research direction, therefore, is to enable the matching step without storing point descriptors. Existing geometry-only methods \cite{zhou2022geometry, wang2024dgc} pursue this idea by relying on geometric cues for point matching, thereby avoiding descriptor storage altogether. In our experiments, however, these existing approaches exhibited a substantial performance gap compared to descriptor-based matching. Bridging this gap remains an open challenge, and advancing geometry-only methods could ultimately enable a fully descriptor-free, structure-based localisation pipeline.

\section{Conclusion}
In this work, we introduce a novel paradigm for enabling structure-based localisation in large scenes by reducing the search space for finding 2D-3D correspondences. We propose visible structure retrieval that replaces search heuristics or image retrieval in traditional pipelines, enabling a retrieval framework that implicitly models the scene instead of explicitly storing observations of it. We show how visible structure retrieval can be formualated as a generative model, and trained in a variational autoencoder setup. We demonstrate that our approach can achieve a localisation performance comparable to the state of the art, while requiring lower computational and storage requirements.

\section*{Acknowledgement}
This work was partially supported by the Wallenberg AI, Autonomous Systems and Software Program (WASP) funded by the Knut and Alice Wallenberg Foundation.

\bibliography{egbib}
\end{document}